\documentclass[final]{cvpr}

\usepackage{times}
\usepackage{epsfig}
\usepackage{graphicx}
\usepackage{amsmath}
\usepackage{amssymb}
\DeclareUnicodeCharacter{0306}{}
\usepackage[pagebackref=true,breaklinks=true,colorlinks,bookmarks=false]{hyperref}

\def\cvprPaperID{****} % *** Enter the CVPR Paper ID here
\def\confYear{CVPR 2021}

\begin{document}

%%%%%%%%% TITLE
\title{MSN: Efficient Online Mask Selection Network \\for Video Instance Segmentation
}

\author{Vidit Goel\textsuperscript{1,2,4}, Jiachen Li\textsuperscript{1,3}\thanks{Equal contribution}, Shubhika Garg\textsuperscript{2}\footnotemark[1]  , Harsh Maheshwari\textsuperscript{2}, Humphrey Shi\textsuperscript{1,3,4} \\
\small \textsuperscript{1}SHI Lab $@$ University of Oregon,
\textsuperscript{2}Indian Institute of Technology Kharagpur,
\textsuperscript{3}University of Illinois at Urbana-Champaign,\\ \small \textsuperscript{4}Picsart AI Research (\textbf{PAIR})\\
}

\maketitle
% \author{Vidit Goel 
% IIT Kharagpur  UIUC PAIR\\
% \and

%%%%%%%%% ABSTRACT
\begin{abstract}
  In this work we present a novel solution for Video Instance Segmentation(VIS), that is automatically generating instance level segmentation masks along with object class and tracking them in a video. 
  %We improve and adapt the mask selection based  method FrameSelect for VIS. 
  Our method improves the masks from segmentation and propagation branches in an online manner using the \textbf{Mask Selection Network} (MSN) hence limiting the noise accumulation during mask tracking. We propose an effective design of MSN by using patch-based convolutional neural network. The network is able to distinguish between very subtle differences between the masks and choose the better masks out of the associated masks accurately. Further we make use of temporal consistency and process the video sequences in both forward and reverse manner as a post processing step to recover lost objects. The proposed method can be used to adapt any video object segmentation method for the task of VIS. Our method achieves a score of 49.1 mAP on 2021 YouTube-VIS Challenge and was ranked third place among more than 30 global teams.
  Our code will be available at:  \href{https://github.com/SHI-Labs/Mask-Selection-Networks}{https://github.com/SHI-Labs/Mask-Selection-Networks}.
\end{abstract}

%%%%%%%%% BODY TEXT
\section{Introduction}
Video understanding has gained a lot of attention with the abundance of data as well as methods for analysing and processing videos. In this work, we particularly focus on Video Instance Segmentation. The task here is to perform instance segmentation and classification on each frame and track the instances through out the video. A very similar task is that of unsupervised video object segmentation~\cite{Caelles_arXiv_2019}. The major difference is that there is no need to predict object classes. 
Due to the large similarity between the two tasks we develop our solution based on the best performing method FrameSelect~\cite{garg2021mask}.

The extra challenges that are present in VIS compared to unsupervised video object segmentation are majorly due to classification task. In case of VIS the method is penalized for false positives, hence even if a method produces very good masks if the classification is wrong the score would be low. Due to this only the objects whose classification is score is very good should be added as a true detection in a frame. However, this can lead to under detection of instances. Other problems which are common to both the task are also present, such as occlusion, motion blur, re-identification, presence of multiple objects of similar visual appearances and absence of information of objects that needs to tracked. Keeping the challenges in mind we adapt and improve the FrameSelect~\cite{garg2021mask} method.

Our method consists of three stages 1) instance segmentation 2)  efficient online mask selection and temporal propagation 3) post processing. From the discussion above, it is clear that object classification is very important for good performance. Hence, we develop a strong instance segmentation module based on Swin tranformer ~\cite{liu2021swin} trained on various open source datasets including
ImageNet~\cite{krizhevsky2012imagenet} MS COCO~\cite{lin2014microsoft} and OpenImages~\cite{kuznetsova2020open}. In stage 2, we handle the task related to mask propagation and object addition. In this work we propose Mask selection network (MSN) to choose better masks from the available masks of an instance in an online manner. This limits the noise accumulation at the time of mask propagation. Finally, in the stage 3 we use temporal consistency of a video sequence to handle under detection of objects.
Most of the recent approaches~\cite{lin2021video, gberta_2020_CVPR, luiten2019video} divide the task into instance segmentation and mask propagation which is similar to ours. However, our method is significantly different from them as it provides a mechanism to assess the quality of masks at the time of propagation of masks and choose the best mask from the available masks. Further, we provide an efficient yet effective post processing method.
Our method proposes a general pipeline to adapt any video object segmentation method for the task of VIS.

% \begin{figure*}
% \begin{center}
% \fbox{\rule{0pt}{2in} \rule{.9\linewidth}{0pt}}
% \end{center}
%   \caption{Example of a short caption, which should be centered.}
% \label{fig:short}
% \end{figure*}

%------------------------------------------------------------------------
\section{Related work}

\noindent \textbf{Video Object Segmentation}
In semi-supervised setting the annotations of first frame are given and they need to segment and track those objects throughout the video in a class-agnostic manner.  Many approaches in this field deal with online learning and fine tuning \cite {luiten2018premvos, wang2019object, xu2019mhp} using the ground truth information from the first frame. These types of methods show impressive performance, however they are not suitable for real time video object segmentation as they take a lot of time. Some other variations of method involve mask propagation \cite {perazzi2017learning, li2018instance, oh2019video} using previous frames and feature matching \cite {voigtlaender2019feelvos, oh2018fast, yang2020collaborative, yang2020cfbi+} of the embedding of the current frame with the stored templates. CFBI~\cite{yang2020collaborative} does local feature matching at multiple scales and treats foreground and background as separate classes, hence, giving better results.\\

\noindent \textbf{Video Instance Segmentation} The works in this category can be divided into two categories, i) treats instance segmentation and tracking as two separate tasks~\cite{luiten2019video, wang2019empirical} and ii) has end to end trainable pipeline~\cite{bertasius2020classifying, cao2020sipmask, athar2020stem, lin2020video, yang2019video, lin2021video, fu2020compfeat, fu2021learning}. 
The methods in first category fine-tune each stage individually and are able to achieve good results. For example Ensemble VIS \cite{luiten2019video} has separate modules for classification and segmentation and finally tracks using a ReID module. The methods in second category train their method in end to end manner. Methods such as~\cite{bertasius2020classifying, lin2021video} are based of propagating masks detected in previous frame whereas methods like~\cite{yang2019video, cao2020sipmask, QueryInst} solve the task by first detecting and then tracking the objects. Different from others ~\cite{athar2020stem} treated the video as 3D volumes and track and segment objects in a single stage. Our method falls in the first category.

\noindent \textbf{Instance Segmentation}
Previous instance segmentation methods can be categorized into two groups: two-stage methods like Mask RCNN~\cite{he2017mask} and one-stage methods like TensorMask~\cite{chen2019tensormask, xu2020deep}. For two-stage methods, they follow the idea from two-stage object detector that localize proposals then segment instances from proposals. Mask RCNN adds a segmentation branch along with detection head. Following methods HTC~\cite{chen2019hybrid} and CBNet~\cite{liu2020cbnet} employ multi-stage head for accurate instance segmentation. For one-stage methods, CenterMask~\cite{lee2020centermask} and Yolact~\cite{bolya2019yolact} achieve real-time instance segmentation without generating proposals. SOLO~\cite{wang2020solo} and SOLOv2~\cite{wang2020solov2} further segments each instance from individual feature map. Recently, with the development of vision transformer~\cite{waswani2017attention, hassani2021escaping}, using transformer-based model as backbone like Swin~\cite{liu2021swin} for instance segmentation achieves new state-of-the-art performance on COCO~\cite{lin2014microsoft} benchmark.

%-------------------------------------------------------------------------
\section{Method}
\subsection{Overall framework}
 Our overall framework leverages an effective multi-stage design for video instance segmentation and consists of 3 stages. Its overall pipeline is similar to Frame Select~\cite{garg2021mask} which was proposed for unsupervised video object segmentation~\cite{Caelles_arXiv_2019}.
 In our stage 1, instance segmentation is performed using Swin transformer backbone~\cite{liu2021swin} based cascaded mask R-CNN. In stage 2 we adapt a video object segmentation method for VIS task. Specifically, we used CFBI~\cite{yang2020collaborative} and initialized it using the masks generated in stage 1. The masks are improved in an online manner using a Mask selection network (MSN) which enables to capture very fine details. MSN is able to select better masks even with very subtle differences between the masks. In stage 3, we propose an efficient yet effective post processing pipeline which utilizes temporal consistency in a video in both forward and backward direction recover and track missing objects.
\subsubsection*{Stage 1: Instance segmentation}

 For instance segmentation, we develop our model based on Swin~\cite{liu2021swin} transformer and cascade mask R-CNN with mmdetection~\cite{chen2019mmdetection} toolbox. We use ImageNet~\cite{krizhevsky2012imagenet} pretrained Swin-base model as our backbone and build a cascade mask R-CNN model with 3 stages. The VIS dataset contains 40 classes and we add images from MS COCO~\cite{lin2014microsoft} and OpenImages~\cite{kuznetsova2020open} for data augmentation. Since COCO contains 80 classes and OpenImages contains hundreds of classes, we map similar classes from these two datasets to VIS dataset. We also observe that data distribution is imbalanced in VIS dataset and to solve this issue, we add a data balanced sampler to improve sampling rate of rare classes and improve instance segmentation result. Only the instance masks which have object scores above a certain threshold are passed to stage 2.
 
 \begin{figure*}
\begin{center}
\includegraphics[ width=16.5cm]{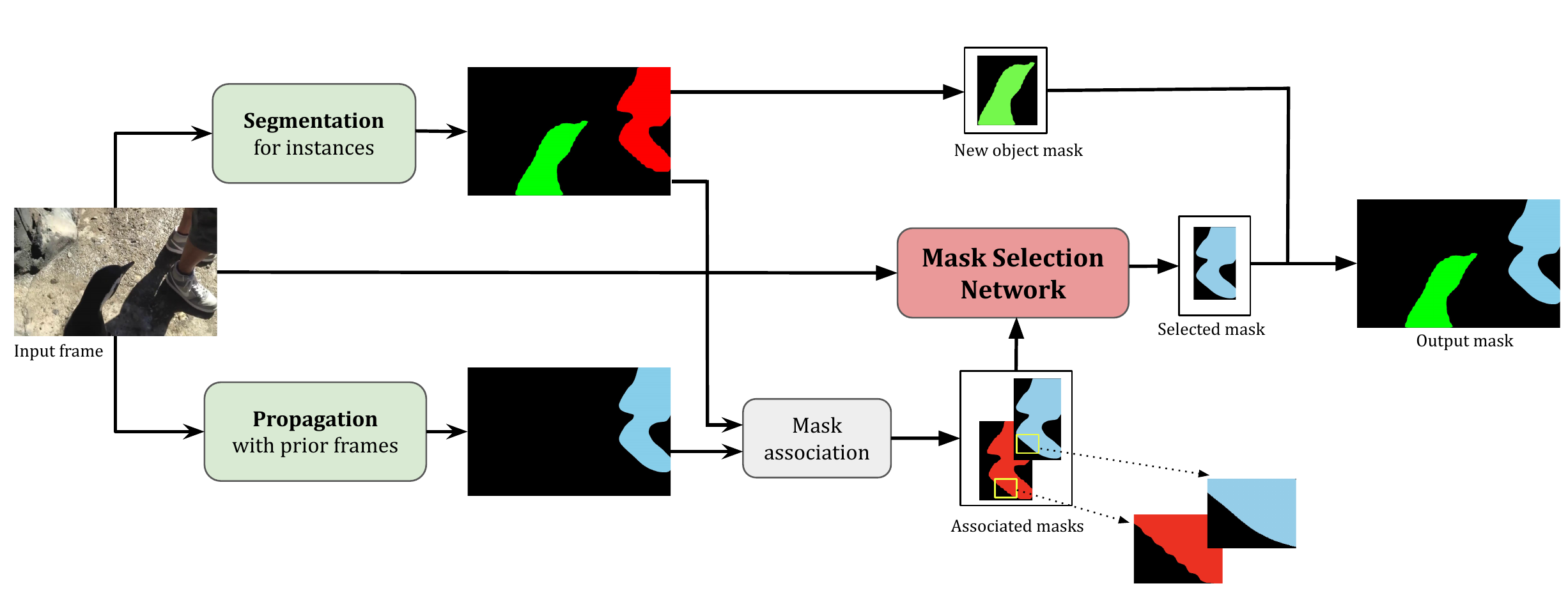}
\end{center}

   \caption{Block diagram of stage 2 of our algorithm. Given an input frame, two sets of mask are generated using segmentation from swin and propagation from CFBI. It is followed by mask association and identifying new objects detected by Swin. The associated pairs are sent to the MSN and best mask is selected to be propagated further. The new objects are added only if it does not intersect with an object of the same class. The outputs from this stage are sent to post processing step in i.e. stage 3.}
\label{fig:block}
\end{figure*}

\subsubsection*{Stage 2: Efficient online mask selection and temporal propagation}

In order to propagate the masks we adapt a video object segmentation method CFBI~\cite{yang2020collaborative} for our task. In case of video object segmentation the annotations of the objects are given when they first appear in a video. In this case, we initialize CFBI by using masks from stage 1 as annotations for first frame. Using CFBI gives 2 major benefits, first one is being able to generate instance masks by incorporating the temporal information and second it helps in tracking. However, there can be cases when the annotations of first frame are noisy which can lead to noise accumulation and improper propagation of masks. There can also be cases when object gets detected in some intermediate frames. In order to handle the mentioned challenges
we propose the following algorithm.

Let  $S_t$ and  $P^i_t$ be the set of segmented masks produced by Swin and propagated mask by CFBI for frame $t$. The two sets of masks are passed to association module, where a bipartite matching is done between the object masks present in both the sets. A 2D matrix is formed whose rows and columns are the objects present in $S_t$ and  $P_t$ respectively and $v_{ij}$ is the IOU between the $i^{th}$ object mask in $S_t$ and $j^{th}$ object mask in $P_t$. The assignment is done using Hungarian algorithm. Object masks in $S_t$ having a IOU higher than 0.1 are associated to corresponding object masks in $P_t$ and the rest of the objects are kept in consideration for new objects. A new object is added only if it does not intersect with an object of the same class. This ensures to limit the addition of false positives. Now for every associated object we have two mask proposal one from Swin and another from CFBI. We use the proposed Mask selection network (explained in Sec. \ref{sn}) for selecting the better mask of the two. This ensures that only the best mask out of the available masks gets propagated hence limiting the noise accumulation. The complete schematic of the pipeline is shown in the Fig. \ref{fig:block}. Note that any other network other than CFBI can also be used. Our method proposes a general pipeline to adapt video object segmentation methods effectively for the task of VIS.

\subsubsection*{Stage 3: Post processing}

\begin{figure}
\centering
\includegraphics[width=8cm]{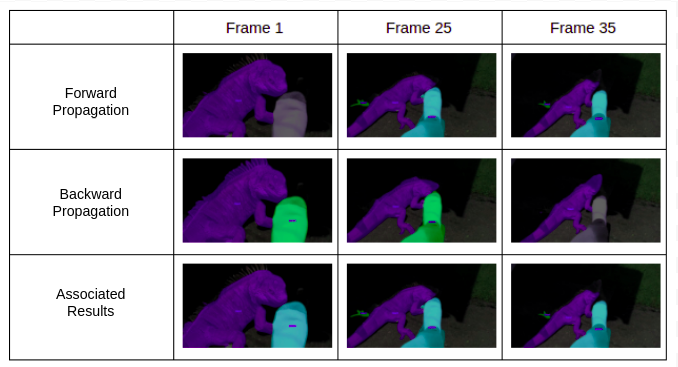}\hspace{10mm}
\vspace{-0.5cm}
\caption{Results for forward and backward association}
\label{fig:forback}
\end{figure}

We leverage the temporal consistency in a video and process the video sequence using stage 2 in both forward and backward manner.
% We run stage 2, two times, that is we have forward and backward implementation. 
In the forward implementation, we start with the instance masks in the first frame of the video sequence and move ahead in time using stage 2 algorithm. In the backward implementation, we start from the instance masks in the last frame and move backward in time using stage 2 algorithm. 
%  This ensures that at
 Hence, at the end of this step, there are 2 outputs for every frame in the video sequence. The next step is to combine these outputs to create a single unified output. To understand the importance of this step, consider a scenario where an object is present in the video from the beginning of the sequence, however that object does not get detected till the $10^{th}$ frame. Hence in the forward implementation, the object will be propagated only from the $10^{th}$ onwards and in the backward implementation, it will only be present in the first $10$ frames. So combining both of them, will ensure that the object is present in the entire video. This is achieved by associating all the objects in every frame using Hungarian algorithm in the forward and backward pass. The sequences of the objects are combined if an object is associated in at least one frame. Having this approach gives us freedom of choosing high threshold value for object score because even if the object is present in few frames it will be tracked through out the video. This helps in reducing false positives. Fig \ref{fig:forback} shows an example of such association.
 
Further we also employ human-object association similar to \cite{yang2019video}. Specifically, we associate boat, motorbike, skateboard, snowboard, surfboard, and tennis racket  with humans. This helps in accurate tracking of objects.
%  Another post processing inspired from \cite{yang2019video},  includes associating human-centric object such as boat, motorbike, skateboard, snowboard, surfboard, and tennis racket  with humans. This association of such objects to human helps improve tracking. 

\subsection{Mask selection network}
\label{sn}

\begin{figure}
\centering
\includegraphics[width=8cm]{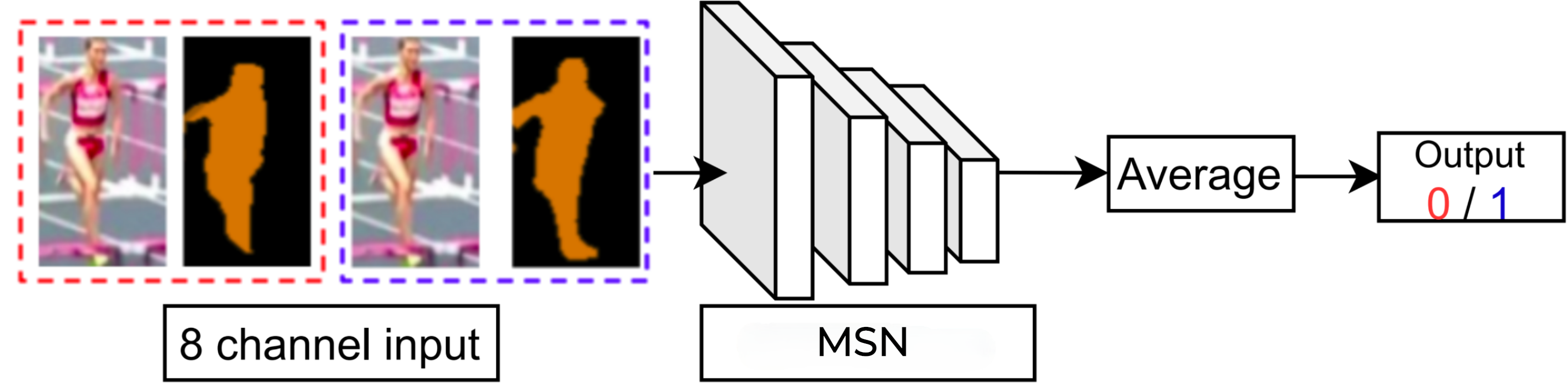}\hspace{10mm}
\vspace{-0.5cm}
\caption{Mask selection network  for selecting better mask of the two input masks.}
\label{fig:sn}
\end{figure}

In this work we propose an improved version of \textit{selector net}~\cite{garg2021mask} named Mask selection network (MSN). The network proposed in ~\cite{garg2021mask} consists of a feature extractor backbone and the feature vectors are aggregated in spatial dimension before feeding them to fully connected layers. This architecture leads to the loss of spatial information due to feature aggregation in the spatial dimensions. Further, the network was trained with the task of predicting scores of the two masks. However, the task of selecting a better mask is more similar to the task of discriminator in GAN. Given two masks, the network needs to tell which mask is better than the other mask. Hence, for this task a relativistic discriminator~\cite{wang2018esrgan} based architecture would be more suitable.

\begin{figure*}
\begin{center}
\includegraphics[width=16cm]{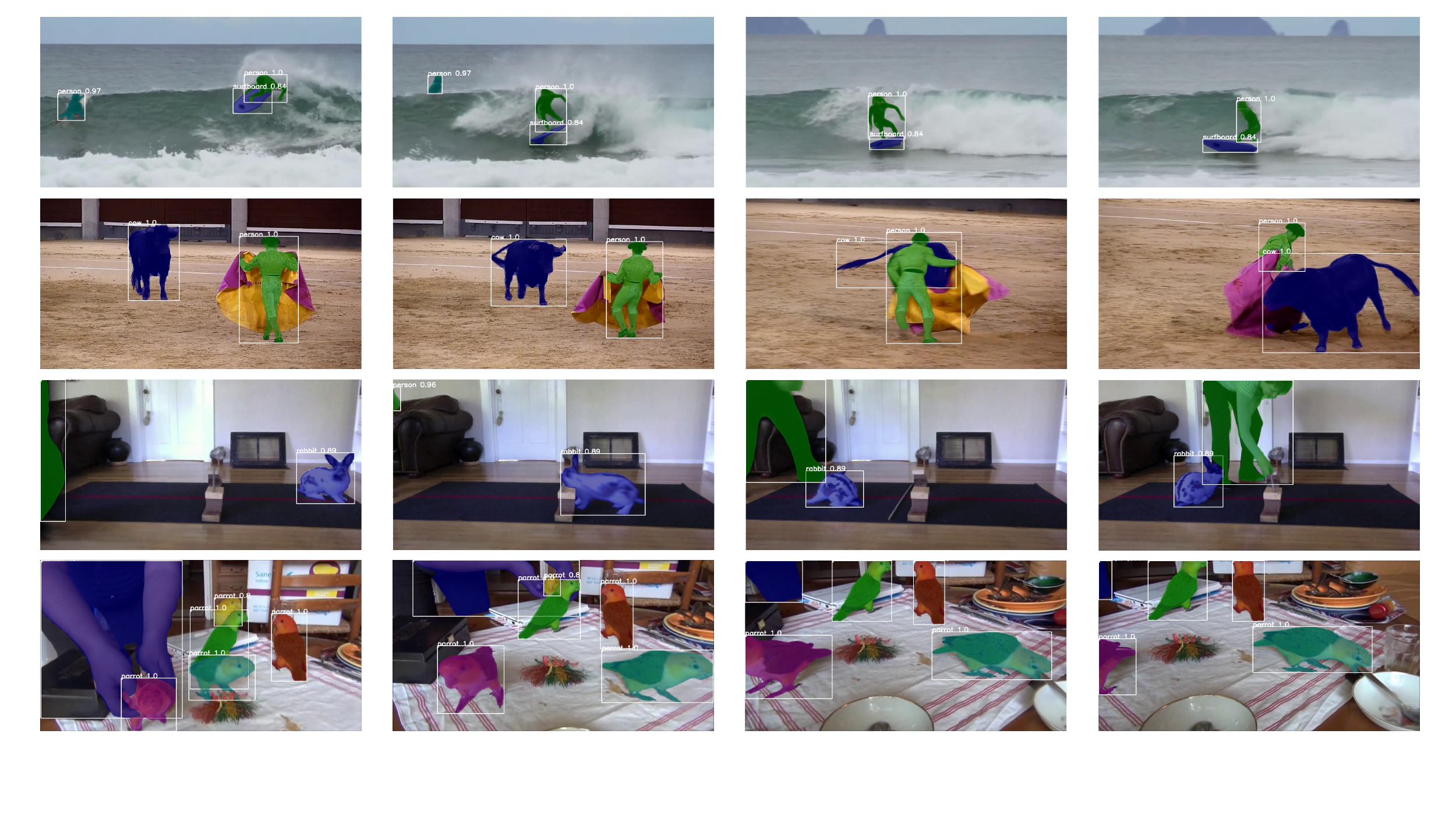}
\end{center}
   \vspace{-1cm}
   \caption{Qualitative results on Youtube Video Instance Segmentation 2021 test dataset.}
\label{fig:2}
\end{figure*}

We use a patch-based network~\cite{isola2017image} because, in a candidate mask, some region could be better and other much worse compared to other mask in input pair. For each patch, the network outputs a score indicating which of the masks in the input pair is better. Finally, the information from each patch is aggregated to make the final decision. The schematic representing the MSN is shown in Fig. \ref{fig:sn}. The data preparation is similar to that in FrameSelect~\cite{garg2021mask}.

The input to MSN consists of 2 binary masks and RGB image corresponding to both the masks, hence the input consists of 8 channels. A tight bounding box around the mask is cropped from the complete image and is used as input as shown in Fig. \ref{fig:sn}.  This is different from ~\cite{garg2021mask} where the complete image is used as input. The images are resized to $256 \times 256$ before feeding to the network. MSN is trained using binary cross entropy loss. The new network  is able to achieve an accuracy of $74\%$ on the validation split whereas the one proposed in \cite{garg2021mask} achieved $69\%$. Note, both the networks were trained on the same training dataset generated using youtube training dataset~\cite{yang2019video} for fair comparison between them. MSN is more accurate as well as efficient compared to the \textit{selector net}~\cite{garg2021mask} with only 2.22 GMac operations compared to 19.45 GMac operations. Table 1. summarizes the comparision between MSN and selector network~\cite{garg2021mask}.

\section{Experiments}

\begin{table}[]
\begin{tabular}{llll}
\hline
Network      & Acc. \% & FLOPs (GMac) & Params(M) \\ \hline
Selector Net~\cite{garg2021mask} & 69          & 19.45        & 11.7       \\ \hline
\textbf{MSN}       & \textbf{74}          & \textbf{2.22 }        & \textbf{11.1}       \\ \hline

\end{tabular}

\caption{Comparison between MSN and selector net~\cite{garg2021mask} in terms of accuracy and compute complexity}
\label{tab:sn}
\end{table}

\subsection{Implementation details}
\noindent \textit{Swin-based Instance Segmentation} For training instance segmentation model, we use Swin-base as our backbone and cascade mask R-CNN as instance segmentation model. For data augmentation, we make class project map and convert Openimage and MS COCO data to classes in VIS dataset and pretrain our model on these open-sourced data. Then, we add a balanced data sampler to improve sampling rate of rare classes and a ViT-based strong classifier to re-classify wrong classifications from the Swin model. The whole network is built based open-sourced implementation of Swin model and trained on 8 RTX 2080Ti GPU. The embedding channel of Swin-base model is 128 and the number of cascade mask R-CNN head is 3. The total batch size is 16 with 2 on each GPU and learning rate is 0.0001. We use Adamw as optimizer with weight decay at 0.05 and warm-up training with 1000 iterations. The total training epoch is 6 with learning rate decay at epoch 3 and 5. For data augmentation in training, we use multi-scale training scales at $[(1333, 800)$, $(1333, 768)$, $(1333, 736)$, $(1333, 704)$, $(1333, 672)$, $(1333, 640)]$ and random flip with probability at 0.5. During inference, we tried different testing test augmentations including multi-scale inference and image flip but they didn't bring further improvements based on our experiments. More experimental results are introduced and shared in section 4.2\\

\textit{Mask selection network} was trained using the data generated from  youtube training dataset~\cite{yang2019video}. The resulting dataset consisted of ~80k training images and ~8k validation images. Network was trained using a batch size of 512 and learing rate of 0.01 using Adam optimizer. The network was trained for 40 epochs and learning rate was scaled by a factor 0.1 after every 10 epochs. The input pair of masks were randomly shuffled at the time of training to prevent the network getting biased towards the ordering of masks. The training was conducted on 8 RTX 2080 GPU.

We used the publicly available implementation of CFBI~\cite{yang2020collaborative}. At the time of inference we used multi-scale and flip strategy in CFBI. Specifically, we apply the scales of \{1.0, 1.15, 1.3, 1.5\}. Further, we limited the maximum number of objects in a video to be 15.

\subsection{Results and analysis}

\begin{table}[]

\resizebox{\columnwidth}{!}{
\begin{tabular}{|l|l|l|l|l|l|}
\hline
\multicolumn{1}{|c|}{Team} & \multicolumn{1}{c|}{mAP} & \multicolumn{2}{c|}{AP}                           & \multicolumn{2}{c|}{AR} \\ \hline
\multicolumn{1}{|c|}{}     & \multicolumn{1}{c|}{}    & \multicolumn{1}{c|}{50} & \multicolumn{1}{c|}{75} & 1          & 10         \\ \hline
tuantng                    & 54.1                    & 74.2                   & 61.6                   & 43.3      & 58.9      \\ \hline
eastonssy                  & 52.3                    & 76.7                   & 57.7                   & 43.9      & 57.0      \\ \hline
\textbf{Ours }                   & \textbf{49.1}                    &\textbf{ 68.1 }                  &\textbf{ 54.5 }                  & \textbf{41.0 }     & \textbf{55.0}      \\ \hline
linhj                      & 47.8                    & 69.3                   & 52.7                   & 42.2      & 59.1      \\ \hline
hongsong.wang              & 47.6                    & 68.4                   & 52.9                   & 41.4      & 54.6      \\ \hline
gb7                        & 47.3                    & 66.5                   & 51.1                   & 40.5      & 51.6      \\ \hline
zfonemore                  & 46.1                    & 64.4                   & 51.0                   & 38.3      & 50.6      \\ \hline
DeepBlueAI                 & 46.0                    & 64.6                   & 52.0                   & 38.7      & 54.2      \\ \hline
\end{tabular}}
\caption{Results on Youtube VIS 2021 test dataset compared to other participants in the challenge.}
\label{tab1}
\end{table}

The proposed method is tested on the Youtube VIS test dataset which contains 453 video sequences. Further, extensive ablation studies were conducted on the validation split of the dataset which contains 421 videos. Table \ref{tab1} shows our results in comparision to top 8 teams. It can be observed that although our AP50 ranks fifth, still we are able to achieve third position. This implies that our tracking and mask quality is more consistent  across various IOU thresholds compared to other method which can be attributed to our online mask improvement using MSN.

In order to understand the importance of various tricks applied in instance segmentation we did an ablation study for each component. The results are shown in Table 2. Other parameters for CFBI such as multiscale inferencing are kept same. From the results it is evident that a strong instance segmentation method is crucial for video instance segmentation. Similar conclusion can be drawn from the work~\cite{QueryInst}, in which they simply improve upon the instance segmentation pipeline to get better results.

\begin{table}[]
\centering

\begin{tabular}{ll}
\hline
\textbf{Model description }              & \textbf{mAP}  \\ \hline

COCO Pretrained + VIS finetune  & 41.3 \\ \hline
COCO + VIS trained              & 47.1 \\ \hline
COCO + VIS + OpenImages trained & 48.3 \\ \hline
+ Balanced sampling             & 48.8 \\ \hline
\end{tabular}
\caption{Ablation study on instance segmentation (stage 1). Experiments are done on Youtube VIS 2021 validation dataset.}
\label{tab2}
\end{table}

\begin{table}[]
\centering

\begin{tabular}{ll}
\hline
\textbf{Method}                   & $\Delta$ mAP \\ \hline
Mask selection network & + 2.6 \\ \hline
Multi scale testing(CFBI) &  + 1.0 \\ \hline
Forward/Backward         & + 0.8 \\ \hline
Object addition criteria & + 0.2 \\ \hline
Object Association       & + 0.2 \\ \hline
\end{tabular}
\caption{Ablation study on various tricks in stage 2 and 3. Here we state the average improvement in mAP across various experiments on Youtube VIS 2021 validation dataset.}
\label{tab:cfbi}
\end{table}

Further, we also studied the improvements brought by various tricks namely MSN, multiscale inferencing in CFBI~\cite{yang2020collaborative}, post processing in stage 3 and object intersection criteria while adding new objects in stage 2 and results are shown in the Table \ref{tab:cfbi}. It can observed that MSN contributes highest in improving results.

\section{Conclusion}

In this work we propose a novel approach for adapting any video object segmentation methods for video instance segmentation. We improve upon the mask selection based method by proposing a mask selection network which can accurately choose a better mask of the given two masks. The mask selector network is highly computationally efficient compared to selector net and is more accurate. The proposed method has a modular structure which will improve as each modules become more efficient. We also propose a post processing pipeline which employs forward and backward processing of video to recover lost objects. 
We believe that instance segmentation is a crucial task in VIS and in order to make significant progress in the task a strong instance segmentation pipeline which incorporates temporal information is necessary.

{\small
\bibliographystyle{ieee_fullname}
\bibliography{cvpr}
}

\end{document}